\documentclass[10pt,twocolumn,letterpaper]{article}

\usepackage{iccv}
\usepackage{times}
\usepackage{epsfig}
\usepackage{graphicx}
\usepackage{amsmath}
\usepackage{amssymb}

\usepackage[accsupp]{axessibility}
\usepackage{bm}
\usepackage{bbm}
\usepackage{booktabs}

\usepackage{algorithm}
\usepackage{algpseudocode}

\usepackage[breaklinks=true,bookmarks=false]{hyperref}

\iccvfinalcopy 

\def\iccvPaperID{****} 

\ificcvfinal\fi

\makeatletter
\newcommand{\printfnsymbol}[1]{%
  \textsuperscript{\@fnsymbol{#1}}%
}
\makeatother

\begin{document}

\title{Distilling Holistic Knowledge with Graph Neural Networks}

\author{Sheng Zhou\textsuperscript{1,2}\thanks{Equal Contribution}, Yucheng Wang\textsuperscript{1}\printfnsymbol{1}, Defang Chen\textsuperscript{1},Jiawei Chen\textsuperscript{3},Xin Wang\textsuperscript{4},Can Wang\textsuperscript{1}, Jiajun Bu\textsuperscript{1}\thanks{Corresponding Author}\\
\textsuperscript{\rm 1}Zhejiang Provincial Key Laboratory of Service Robot, Zhejiang University\\
\textsuperscript{\rm 2}School of Software Technology, Zhejiang University
\\
\textsuperscript{\rm 3}University of Science and Technology of China
\textsuperscript{\rm 4}Tsinghua University
\\
{\tt\small \{zhousheng\_zju,wangyuc,defchen\}@zju.edu.cn, cjwustc@ustc.edu.cn, xin\_wang@tsinghua.edu.cn}
\\
{\tt\small \{wcan,bjj\}@zju.edu.cn}
}
\maketitle
\ificcvfinal\thispagestyle{empty}\fi

\begin{abstract}
   Knowledge Distillation (KD) aims at transferring knowledge from a larger well-optimized teacher network to a smaller learnable student network.
   Existing KD methods have mainly considered two types of knowledge, namely the individual knowledge and the relational knowledge.
   However, these two types of knowledge are usually modeled independently while the inherent correlations between them are largely ignored.
   It is critical for sufficient student network learning to integrate both individual knowledge and relational knowledge while reserving their inherent correlation.
   In this paper, we propose to distill the novel holistic knowledge based on an attributed graph constructed among instances. The holistic knowledge is represented as a unified graph-based embedding by aggregating individual knowledge from relational neighborhood samples with graph neural networks, the student network is learned by distilling the holistic knowledge in a contrastive manner.
   Extensive experiments and ablation studies are conducted on benchmark datasets, the results demonstrate the effectiveness of the proposed method.
   The code has been published in \url{https://github.com/wyc-ruiker/HKD}
   \end{abstract}

\section{Introduction}
\label{sec:intro}

\noindent 
Deep Neural Networks (DNNs) have shown tremendous success in various applications \cite{he2016deep,silver2017mastering,hamilton2017inductive,sandler2018mobilenetv2,devlin2019bert, zhang2017active}. However, their success heavily relies on extensive computational and storage resources, which are usually unavailable in embedded and mobile systems.
To reduce the cost while maintaining satisfactory, knowledge distillation \cite{hinton2015distilling} is proposed to transfer knowledge from a larger well-trained \textit{teacher network} to a smaller learnable \textit{student network}, hoping that the transferred knowledge will benefit the student network.

\begin{figure}
    \centering
    \includegraphics[width=0.45\textwidth]{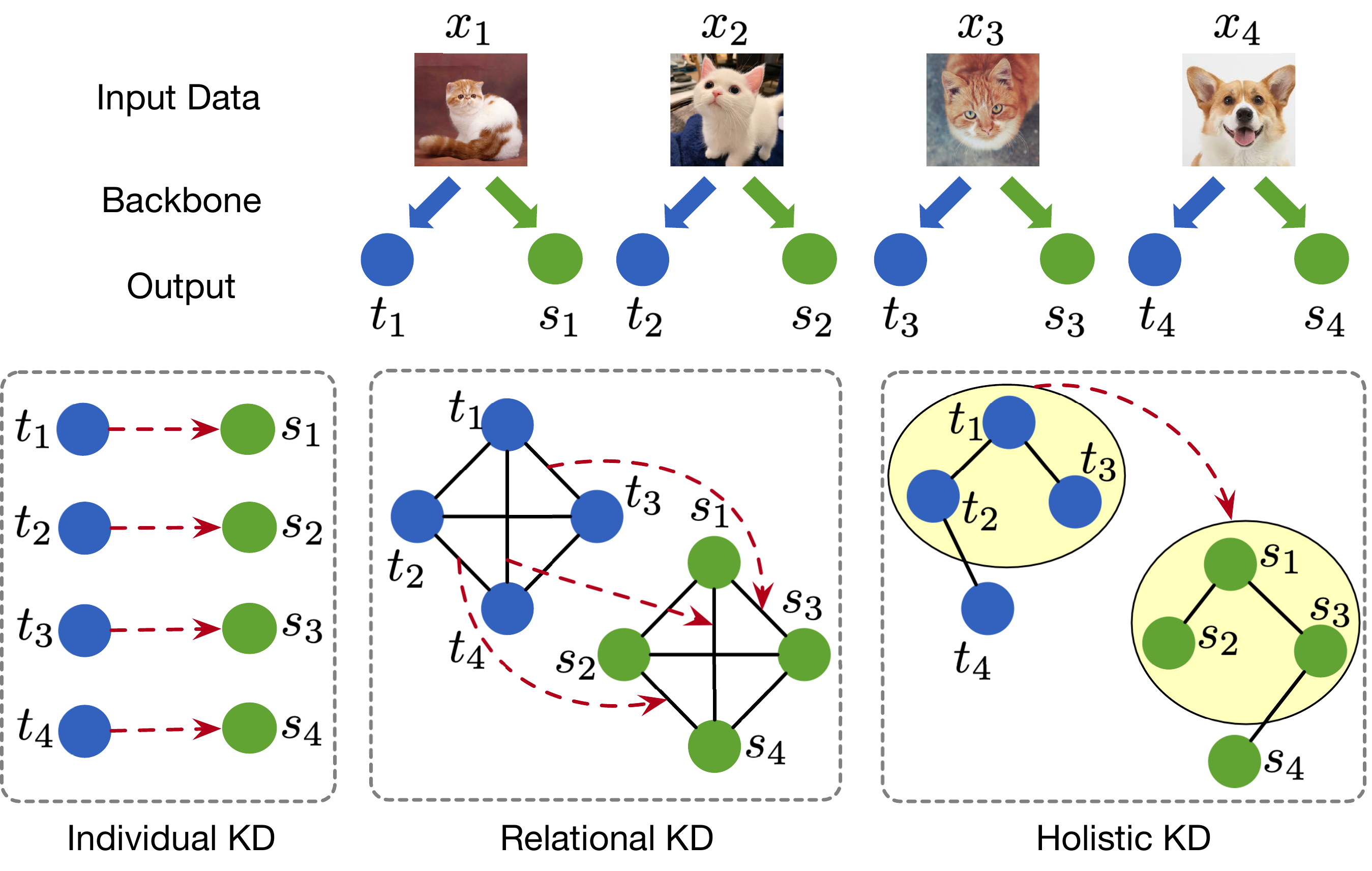}
    \caption{Comparison between Individual/Relational/Holistic Knowledge Distillation. The blue circle indicates the teacher representation, and the green circle indicates the student representation. 
    The red arrow denotes the knowledge transfer from the teacher network to the student network. The yellow area in the holistic KD indicates the unified graph-based representation. }
    \label{fig:intro}
\end{figure}

The knowledge distilled from the teacher network has played the central role in knowledge distillation.
Among existing knowledge distillation methods, two types of knowledge have been widely studied, namely the \textit{individual knowledge} and the \textit{relational knowledge}.
The individual knowledge is extracted from each data instance independently and provides more favorable supervision than the discrete labels, including logits \cite{hinton2015distilling}, feature representations \cite{tian2019contrastive,passalis2018learning} and feature maps \cite{romero2014fitnets,zagoruyko2016paying,li2020local}, etc.
The relational knowledge \cite{peng2019correlation,liu2019knowledge,park2019relational,lee2019graph} is extracted from pairs of instances which is invariant to the difference between architectures of the teacher network and the student network.

Despite the success of the above two types of knowledge, existing methods have extracted them independently, ignoring their inherent correlations.
However, each type of knowledge that extracted independently will be insufficient for the student network learning, especially when the capability of the teacher network is limited.
Intuitively, the individual knowledge and the relational knowledge can be treated as two views of the same teacher network, which are naturally correlated. 
The closely related instances tend to have similar individual features and shared patterns, which is critical for more discriminative student network learning.
Simultaneously integrating the individual and relational knowledge while reserving their inherent correlation is of primal importance for knowledge distillation.

To resolve the above limitations, we propose the \textbf{H}olistic \textbf{K}nowledge \textbf{D}istillation (HKD) method with graph neural networks.
We introduce a novel holistic knowledge which is an integration of both individual knowledge and relational knowledge.
Given the feature representations and predictions learned by the teacher and the student network, we first build an attributed graphs for each network, where each node denotes an instance, the node attributes denote the learned feature representation, the edges among instances are constructed by the K-nearest-neighbor (KNN) on the predictions.
Inspired by the recent success of Graph Neural Networks (GNNs) \cite{hamilton2017inductive,lee2019graph} in simultaneously modeling network topology and node attributes, we extract the holistic knowledge by aggregating node attributes from the neighborhood samples in the attributed graph, represented as a unified graph-based embedding.
Figure \ref{fig:intro} illustrates the comparison among the individual, relational and holistic knowledge.
We also theoretically prove that existing individual knowledge and relational knowledge are special cases of holistic knowledge under certain conditions.

Given the holistic knowledge represented by graph-based embedding, a naive way of knowledge distillation is directly aligning the embedding of the same instance from the teacher and the student network. 
However, since the student network usually has lower capability than the teacher network, force aligning the graph-based embedding is too strict for transferring the shared patterns in the neighborhood and holistic knowledge.
Instead, HKD aims at maximizing the mutual information between the graph-based representation from the teacher and the student network, which is optimized with InfoNCE estimator\cite{oord2018representation} in a contrastive manner.
The holistic knowledge guides the student network learning in two ways: first, the student should learn similar instance features and relational neighborhood as the teacher network; second, the student should capture similar patterns from the neighborhood instances in the attributed graph.
The memory bank technique is also employed to further improve the training efficiency.
To conclude, we summarize our contributions as follows:
\begin{enumerate}
    \item We propose Holistic Knowledge Distillation (HKD), a novel method to efficiently distill holistic knowledge for the student network learning. 
    \item The proposed HKD method employs graph neural networks to simultaneously  integrate both the individual and relational knowledge into a unified representation, where their inherent relationship can be reserved. 
    \item We conduct extensive experiments on benchmark datasets to evaluate the performance of HKD and motivation of holistic knowledge, the results demonstrates the effectiveness of the proposed HKD method.
\end{enumerate}

\section{Related Work}
\label{sec:related}
\textbf{Knowledge Distillation.} 
Knowledge distillation was first introduced as a neural network compression technique that minimizes the KL-divergence between the output logits of teacher and student networks \cite{ba2014deep,hinton2015distilling}. Compared with discrete labels, the relative probabilities predicted by the teacher network tend to encode semantic similarities among categories, which are important for the student network learning \cite{hinton2015distilling}.
Some subsequent works have been proposed to widen its applicability, such as adding regularization on the logits \cite{wen2019preparing,chen2020online}, intermediate layers \cite{romero2014fitnets,zagoruyko2016paying,chen2020cross,li2020local} or distillation process \cite{yang2019training,yang2019snapshot}.

However, the above mentioned methods distill knowledge contained in each instance independently but ignore the relationship among instances, which is critical to achieve a robust and general student model.
To make up this shortcoming, relational knowledge distillation \cite{park2019relational} is proposed by distilling both instance-wise and relation-wise knowledge.
Given particular layer $l$, GKD \cite{lassance2020deep} build a KNN-based graph on the cosine similarity of the inner representation and the weights represent the strength of proximity between two instances.
However, it requires the layer number of the teacher and student network are same, which is not always satisfied. 
IRG \cite{liu2019knowledge} is then proposed by introducing feature space transformation across layers.
In MHGD \cite{lee2019graph}, the relation-level knowledge is distilled to a graph using an attention network and optimized by minimizing the KL-divergence between the embedded the teacher and student graphs.
Recent works \cite{tian2019contrastive,xu2020knowledge} have incorporated contrastive learning and achieve inspiring results.
CRD \cite{tian2019contrastive} performs contrastive learning by maximizing the mutual information between the teacher and student networks.
SSKD \cite{xu2020knowledge} performs contrastive learning separately in the teacher and student networks, then the model is optimized by minimizing the loss between the output of self-supervised module from two networks. 
To clearly show the most critical contribution of our method, we do not utilize the intermediate information and compare with those methods relying on it in the experimental part.

\textbf{Graph Neural Network.}
Graph Neural Networks (GNNs) \cite{kipf2016semi,hamilton2017inductive} aim at learning node representation by collectively aggregate information from neighborhood instances in graph structure data.
The learned representation can model individual features as well as relationship between instances which is critical for data understanding. 
Profit from this property, GNNs have made remarkable advancements in a great many learning tasks beyond network/graph representation \cite{zhou2018graph,zhou2018prre,zhou2020dge}, including computer vision \cite{garcia2017few,lee2018multi}, natural language processing \cite{rahimi2018semi,beck2018graph} and recommendations\cite{chen2019samwalker,chen2018modeling} etc.
Although success in other domains, to the best of our knowledge, GNNs has not be explored to knowledge distillation and we are the first one to do so.

\section{Preliminaries}
\label{sec:prel}
\subsection{Background and Notations}
Given a dataset $\mathcal{X} = \{\mathbf{x}_{1},\mathbf{x}_{2},\cdots, \mathbf{x}_{N}\}$ from K categories with corresponding labels $\mathcal{Y} = \{\mathbf{y}_{1},\mathbf{y}_{2},\cdots, \mathbf{y}_{N}\}$, where $N$ represents the number of samples in the dataset.
We refer a well-optimized deep neural network with fixed parameters $\mathbf{W}^{t}$ as the teacher network and a relatively shallow neural network with trainable parameters $\mathbf{W}^{s}$ as the student network \cite{hinton2015distilling}.
The feature representations learned by the teacher and student networks are denoted as $\mathbf{f}^{t}\in \mathbb{R}^{d^{t}}$ and $\mathbf{f}^{s}\in \mathbb{R}^{d^{s}}$, which are mainly used in relational knowledge distillation.
It is worth noting that $d^{t}$ and $d^{s}$ may be different especially when the teacher and the student network architectures are different.
The logits predicted by the teacher and student networks are denoted as $\mathbf{z}^{t}$ and $\mathbf{z}^{s}$, which are mainly used in individual knowledge distillation.

\subsection{Vanilla Knowledge Distillation}
The general idea of vanilla knowledge distillation is to distill knowledge from soft targets predicted by the teacher network \cite{hinton2015distilling}.
The soft targets are produced by Softmax function with temperature scaling:
\begin{equation}
    p_{i}(\mathbf{z}; \tau)=\operatorname{Softmax}(\mathbf{z}; \tau)=\frac{e^{z_{i} / \tau}}{\sum^{K}_{k=1} e^{z_{k}/ \tau}}
\end{equation}
where $z^{i}$ is the corresponding logit of the $i$-th class and temperature $\tau$ is normally set to 1. 
Using a higher value for $\tau$ will produce a softer probability distribution over classes.
The student network is then optimized by minimizing the Kullback-Leibler (KL) divergence between soft targets $\mathbf{p}^{t}$ and $\mathbf{p}^{s}$ produced by the teacher and the student networks:
\begin{equation}
    \mathcal{L}_{KD}(\boldsymbol{p}^{s},\boldsymbol{p}^{t})=\frac{1}{N} \sum_{i=1}^{N} \text{KL}\left(\boldsymbol{p}^{s}, \boldsymbol{p}^{t}\right)
\label{equa:kd}
\end{equation}
In vanilla KD, the student network is also trained with the hard labels and the total loss can be formalized as:
\begin{equation}
    \mathcal{L}=\mathcal{L}_{CE}(\boldsymbol{p}^{s},\mathbf{y})+\lambda \mathcal{L}_{KD}(\boldsymbol{p}^{s},\boldsymbol{p}^{t})
\end{equation}
where $\lambda$ is a balancing weight. 
$\mathcal{L}_{CE}$ is the Cross-Entropy (CE) loss between the hard labels and prediction.

\section{Model}
\label{sec:model}
As mentioned earlier, holistic knowledge is expected to integrate both individual knowledge and relational knowledge. 
Inspired by recent success of graph neural networks in simultaneously modeling the network topology and node attributes, we utilize graph neural networks to extract holistic knowledge from the teacher network. 
In the following subsection, we will elaborate on the details of the proposed holistic knowledge distillation (HKD) method.

\begin{figure*}
    \centering
    \includegraphics[width=0.8\textwidth]{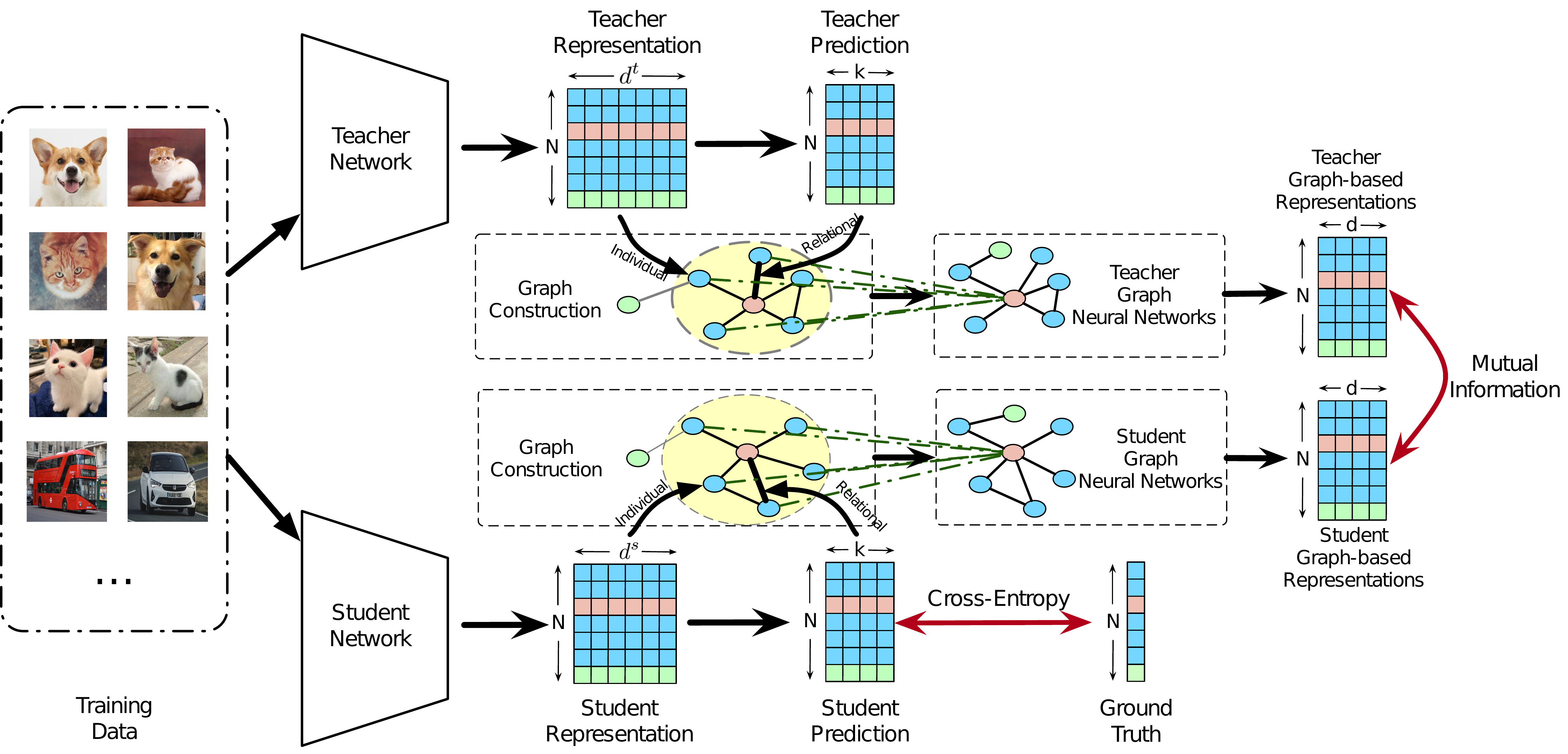}
    \caption{The overall framework of the HKD Method. Three major components are carefully designed: graph construction, graph neural networks, and mutual information estimation to represent, define, and distill the holistic knowledge. The student model is trained under the guidance of ground truth labels and mutual information of the holistic knowledge.}
\end{figure*}

\subsection{Attributed Context Graph Construction}
Given a batch of instances, we first feed them into the teacher network and the student network to get the feature representations $\mathbf{f}^{t},\mathbf{f}^{s}$ as well as prediction $\mathbf{p}^{t},\mathbf{p}^{s}$.
Then we build two attributed graphs $\mathbf{G}^{t} = \{\mathbf{A}^{t},\mathbf{F}^{t}\}$ and $\mathbf{G}^{s}= \{\mathbf{A}^{s},\mathbf{F}^{s}\}$ for the teacher network and the student network,
where $\mathbf{F}^{t}\in \mathbb{R}^{N\times d^{t}},\mathbf{F}^{s}\in \mathbb{R}^{N\times d^{s}}$ are the attributes of nodes in the graph, here we directly use the feature representations learned by the teacher and the student networks; 
$\mathbf{A}^{t},\mathbf{A}^{s}$ are the adjacent matrices of the attributed graphs which are based on the prediction $\mathbf{p}^{t}, \mathbf{p}^{s}$ predicted by the teacher and the student networks:
\begin{equation}
    \mathbf{A}^{t} = \phi(\mathbf{p}^{t}), \quad \mathbf{A}^{s} = \phi(\mathbf{p}^{s})
\end{equation}
where $\phi(\cdot)$ is the KNN-based graph construction function.
Note that the graph $\mathbf{G}^{t}$ is fixed since the teacher network has been well optimized while the graph $\mathbf{G}^{s}$ will be updated during training in both node attributes and graph topology.

The attributed graph defined above enjoys the following properties:
First, compared with fully connected graph among instances built by existing relational knowledge distillation methods, the KNN graph will filter out the most incorrelated sample pairs. 
This is particular important since only a few samples are correlated in randomly sampled batches and provide sufficient information for the node representation learning.
Second, the graph is able to model the inter-class and intra-class information since the edges are constructed based on prediction. The samples from two highly correlated classes will have a high probability to form an edge.
Finally, it is straightforward to jointly extract both the individual and relational knowledge from the attributed context graph with graph neural networks.

\subsection{Holistic Knowledge Distillation}
Inspired by the tremendous success of graph neural networks in simultaneously modeling the network topology and node attributes, we apply Topology Adaptive Graph Convolution Network (TAGCN) \cite{du2017topology,kipf2016semi} on the attributed context graphs $\mathbf{G}^{t}$ and $\mathbf{G}^{s}$ to extract the holistic knowledge.
We use the graph-based representations $\mathbf{H}^{t}\in \mathbb{R}^{N\times g^{t}}$ and $\mathbf{H}^{s}\in \mathbb{R}^{N\times g^{s}}$ of the teacher and student networks to denote the holistic knowledge, which can be calculated as:
\begin{equation}
    \mathbf{H}^{t}=\sum_{l=0}^{L} \left(\mathbf{D}_{t}^{-1 / 2} \mathbf{A}^{t} \mathbf{D}_{t}^{-1 / 2}\right)^{l} \mathbf{F}^{t} \mathbf{\Theta}^{t}_{l}
\label{equ:ht}
\end{equation}
\begin{equation}
    \mathbf{H}^{s}=\sum_{l=0}^{L} \left(\mathbf{D}_{s}^{-1 / 2} \mathbf{A}^{s} \mathbf{D}_{s}^{-1 / 2}\right)^{l} \mathbf{F}^{s} \mathbf{\Theta}^{s}_{l}
\label{equ:hs}
\end{equation}
where $g^{t},g^{s}$ are the dimension of the graph-based representation, $\textbf{D}_{t}=\sum_{j}\textbf{A}^{t}_{ij}$ is the diagonal degree matrix of the teacher network and so is the $\textbf{D}_{s}$ matrix, 
$\mathbf{\Theta}^{s}_{l}$ and $\mathbf{\Theta}^{t}_{l}$ are the learnable weights to sum the results of $l$-th hops together and we set $L=1$ here.

A good student network is expected to distill holistic knowledge from the teacher network by learning similar graph based repesentation $\mathbf{H}^{s}$ with $\mathbf{H}^{t}$.
There exists several vector-wise metrics for measuring their alignment, including the Cosine Similarity, Euclidean distance, etc.
However, these metrics are not suitable for holistic knowledge distillation since the teacher and the student networks usually have different network architectures, there exists a gap between the representation capability.
As a result, directly aligning the graph-based representation $\mathbf{H}^{s}$ and $\mathbf{H}^{t}$ of the same instance may be over refine.
To overcome the limitations, we use the Mutual Information (MI) \cite{tschannen2019mutual} to measure the amount of holistic knowledge distilled from the teacher network to the student network.

Assume that we are given a set of training instances $\mathcal{X}$ with an empirical probability distribution $\mathbb{P}$,
after pushing instances through the teacher and the student networks, the graph-based representations will obey the probability distribution $\mathbf{H}^{t} \sim \mathbb{P}^{t}$ and $\mathbf{H}^{s}\sim \mathbb{P}^{s}$.
We wish to train the student network by maximizing the mutual information between the graph-based representations $\mathbf{H}^{t}$ and $\mathbf{H}^{s}$:
\begin{equation}
    \mathop{\mathcal{L}_{HOL}}_{\mathbf{W}^{s},\mathbf{\Theta}^{t},\mathbf{\Theta}^{s}}=-\mathbf{I}(\mathbf{H}^{t},\mathbf{H}^{s}) 
\end{equation}
where $\mathbf{I}(\cdot)$ denotes the mutual information between two random variables.
Inspired by recent success in mutual information estimation, we use the InfoNCE estimator \cite{oord2018representation} to measure the mutual information, which is defined as:
\begin{equation}
    \mathbf{I}(\mathbf{H}^{t},\mathbf{H}^{s})\geq \mathbb{E}\left[\frac{1}{N}\sum_{i=1}^{N}log\frac{e^{f(\mathbf{h}^{t}_{i},\mathbf{h}^{s}_{i})}}{\frac{1}{N}\sum^{N}_{j=1}e^{f(\mathbf{h}^{t}_{i},\mathbf{h}^{s}_{j})}}\right]
\label{equ:MI}
\end{equation}
where $f(\cdot)$ is the vector-wise similarity function and we use cosine similarity here, $\mathbf{h}^{t}_{i},\mathbf{h}^{s}_{i}$ are the graph-based representations of instance $i$ learned by the teacher network and the student network.
The objective of holistic knowledge distillation can be formulated as:
\begin{equation}
    \mathcal{L} = \mathcal{L}_{CE}+\beta \mathcal{L}_{HOL}
\label{equ:loss}
\end{equation}
where $\beta$ is the weight for linear combination.

\subsection{Efficient Training}
Since the InfoNCE estimator uses all the instances in the dataset as negative samples, computing the holistic knowledge distillation loss with graph neural networks is computational expensive for large scale dataset.
To avoid recomputing the representations for each instance during training, the widely used Memory Bank \cite{wu2018unsupervised} strategy is used for storing them.
However, in the HKD method, the attributed context graph $\mathbf{G}^{t}$ and $\mathbf{G}^{s}$ are constructed on mini-batch with randomly sampled instances.
As a result, the graph-based representations $\mathbf{H}^{t}$ and $\mathbf{H}^{s}$ reflect holistic knowledge in different attributed graphs, which should not be stored in memory bank and serve as negative samples.
To overcome this limitation while improve the efficiency of the HKD method, we maintain two memory banks for the teacher network and the student network, where the feature representation $\mathbf{f}^{t}, \mathbf{f}^{s}$ are stored and serve as the negative samples for training.
The approximate holistic knowledge distillation loss can be formulated as:
\begin{equation}
    \begin{split}
      \widetilde{\mathcal{L}}_{HOL}=\sum^{N}_{i=1}&log\frac{e^{f(\mathbf{h}^{t}_{i},\mathbf{h}^{s}_{i})}}{e^{f(\mathbf{h}^{t}_{i},\mathbf{h}^{s}_{i})}+\sum^{N}_{j=1,j\neq i}e^{f(\mathbf{h}^{t}_{i},\mathbf{f}^{s}_{j})}}\\ +& log\frac{e^{f(\mathbf{h}^{s}_{i},\mathbf{h}^{t}_{i})}}{e^{f(\mathbf{h}^{s}_{i},\mathbf{h}^{t}_{i})}+\sum^{N}_{j=1,j\neq i}e^{f(\mathbf{h}^{s}_{i},\mathbf{f}^{t}_{j})}}  
\end{split}
\label{equ:final_loss}
\end{equation}
The overall framework of the HKD method is illustrated in Algorithm \ref{alg:Framwork}.

\begin{algorithm}[htb] 
 \caption{Holistic Knowledge Distillation.}
 \label{alg:Framwork} 
 \begin{algorithmic}[1] 
  \Require 
  Training dataset $\mathcal{D}=\{(\bm{x}_i, \bm{y}_i)\}_{i=1}^{N}$; A pre-trained teacher model with parameter $\mathbf{W}^t$; A student model with random initialized parameters $\mathbf{W}^s$; 
  \Ensure A well-trained student model; 
  \While{ $\mathbf{W}^s$ is not converged }
  \State 
  Sample a mini-batch $\mathcal{B}$ with size $b$ from $\mathcal{D}$.
  \State
  Forward propagation $\mathcal{B}$ into $\mathbf{W}^t$ and $\mathbf{W}^s$ to obtain feature representation $\mathbf{f}^{t},\mathbf{f}^{s}$ and prediction $\mathbf{p}^{t},\mathbf{p}^{s}$.
  \State
  Construct attributed context graph $\mathbf{G}^{t}$ and $\mathbf{G}^{s}$.
  \State
  Extract holistic knowledge with graph neural networks by Equation (\ref{equ:ht}),(\ref{equ:hs}).
  \State
  Calculate the Mutual information between graph-based representation as Equation (\ref{equ:final_loss}).
  \State
  Update parameters $\mathbf{W}^s$ by backward propagation the gradients of the loss in Equation (\ref{equ:loss}).
  \EndWhile
 \end{algorithmic} 
\end{algorithm}
\subsection{Analysis with Existing Methods}
To further show the generality of HKD, we provide a theoretical analysis that many existing knowledge distillation methods can be viewed as the special cases of our method with certain conditions.

\textbf{Feature-based KD Methods.} 
Feature-based KD methods are popular which only distill the feature representation learned by the teacher network.
Compared with HKD, these methods \cite{tung2019similarity,hinton2015distilling,zagoruyko2016paying,passalis2018learning} ignore the relationship among instances, which can be achieved by setting the $L=0$ or $\mathbf{A} = \text{diag}(N)$ in HKD:
\begin{equation}
    \mathbf{H}^{t} =  \mathbf{F}^{t}\mathbf{\Theta}^{t},\quad \mathbf{H}^{s} = \mathbf{F}^{s}\mathbf{\Theta}^{s}
\end{equation}
where diag($\cdot$) is the diagonal matrix.

\textbf{Relational KD Methods.}
The pairwise relationship of instances captured by these methods \cite{park2019relational,peng2019correlation,tung2019similarity,tian2019contrastive} can be easily reached by setting the feature matrix $\mathbf{H}^{t}, \mathbf{H}^{s} \in \mathbb{R}^{N\times N}$ as the simialrity of feature representation $\mathbf{F}^{t}, \mathbf{F}^{s}$:

\begin{equation}
    \mathbf{H}^{t} = \varphi(\mathbf{F}^{t},\mathbf{F}^{t}),\quad \mathbf{H}^{s} = \varphi(\mathbf{F}^{s},\mathbf{F}^{s})
\end{equation}
where $\varphi(\cdot)$ is the vector-wise similarity function.  
For the methods that do not estimate mutual information, they can be viewed as a special formation of Equation (\ref{equ:MI}) without negative samples.

\section{Experiment}
\label{sec:experiment}
In this section, we first conduct model compression and representation transferability experiments on benchmark datasets to evaluate the proposed HKD method.
Then we conduct several ablation studies on graph construction and graph neural networks to validate their effectiveness.
Finally, we provide the experimental analysis on the hyperparameter sensitivity of the HKD method.

\begin{table*}[!htbp]
    \centering
    \caption{Test accuracy (\%) of the student networks on the CIFAR100 dataset of combining distillation methods with KD. }
    \begin{tabular}{c|ccccc|c}
       \toprule
    \begin{tabular}[c]{@{}c@{}}Teacher\\ Student\end{tabular} & \begin{tabular}[c]{@{}c@{}}ResNet32$\times$4\\ ResNet8$\times$4\end{tabular} & \begin{tabular}[c]{@{}c@{}}ResNet32$\times$4\\ ShuffleNetV2\end{tabular} &  \begin{tabular}[c]{@{}c@{}}VGG13\\ MobileNetV2\end{tabular} & \begin{tabular}[c]{@{}c@{}}ResNet50\\ VGG8\end{tabular} & \begin{tabular}[c]{@{}c@{}}ResNet50\\ MobileNetV2\end{tabular}  & ARI (\%)\\ 
           \midrule

    \begin{tabular}[c]{@{}c@{}}Teacher\\ Student\end{tabular} & \begin{tabular}[c]{@{}c@{}}79.42  \\ 72.79 $\pm$ 0.26  \end{tabular}& \begin{tabular}[c]{@{}c@{}}79.42  \\ 72.63 $\pm$ 0.71  \end{tabular} & \begin{tabular}[c]{@{}c@{}}74.64  \\ 65.33 $\pm$ 0.63  \end{tabular}& \begin{tabular}[c]{@{}c@{}}79.34  \\ 70.56 $\pm$ 0.32  \end{tabular}   & \begin{tabular}[c]{@{}c@{}}79.34  \\ 65.33 $\pm$ 0.63 \end{tabular} & / \\
       \midrule

    KD      & 73.55 $\pm$ 0.20   & 75.38 $\pm$ 0.52     & 68.08  $\pm$ 0.24   & 73.76 $\pm$ 0.09  & 67.83 $\pm$ 0.46 & 126.48 \% \\
    AT+KD   & 74.80 $\pm$ 0.15   & 76.51 $\pm$ 0.16     & 66.37 $\pm$ 0.13    & 73.91 $\pm$ 0.24  & 66.81 $\pm$ 0.11   & 152.84 \%\\
    PKT+KD  & 74.68 $\pm$ 0.07   & 76.16 $\pm$ 0.16      & 68.08 $\pm$ 0.94    & 74.19 $\pm$ 0.27  & 68.42 $\pm$ 0.39  &  55.63 \%\\
    SP+KD   & 73.99 $\pm$ 0.05   & 76.02 $\pm$ 0.34     & 68.46 $\pm$ 0.37    & 73.50 $\pm$ 0.20  & 68.18 $\pm$ 0.57  & 
    80.89 \% \\
    CC+KD   & 74.44 $\pm$ 0.14   & 75.81 $\pm$ 0.20     & 68.54 $\pm$ 0.21    & 73.48 $\pm$ 0.16  & 68.92 $\pm$ 0.16  &
    58.96 \% \\
    RKD+KD  & 74.18 $\pm$ 0.09   & 75.64 $\pm$ 0.24      & 68.24 $\pm$ 0.46    & 73.81 $\pm$ 0.11  & 68.52 $\pm$ 0.14 &   72.15 \% \\
    CRD+KD  & 75.64 $\pm$ 0.25   & 76.41 $\pm$ 0.36     & 69.82 $\pm$ 0.22    & 74.41 $\pm$ 0.31  & 69.86 $\pm$ 0.04 & 
    15.32 \%\\ 
    SSKD+KD  & 75.80 $\pm$ 0.58   & 76.36 $\pm$ 0.38  & 69.12 $\pm$ 0.54    & 74.68 $\pm$ 0.22 &  69.53 $\pm$0.43   & 
    18.86 \%\\        \midrule

    HKD    &  75.63 $\pm$ 0.22    & 76.31  $\pm$  0.30      &69.97 $\pm$ 0.42       &74.86 $\pm$ 0.17   &69.83 $\pm$ 0.15   & 12.94 \%    \\ 
    HKD+KD    & \textbf{76.13 $\pm$ 0.05  } & \textbf{76.92 $\pm$ 0.22  }  & \textbf{70.48 $\pm$ 0.25  }& \textbf{74.88 $\pm$ 0.30}  & \textbf{70.72 $\pm$ 0.32  }& / \\ 
       \bottomrule

    \end{tabular}
    \label{table:class1}
    \end{table*}

\begin{table*}[htpb]
\centering
\caption{Test accuracy (\%) of the student networks on the TinyImageNet dataset of combining distillation methods with KD.}
\begin{tabular}{c|ccccc|c}
\toprule
\begin{tabular}[c]{@{}c@{}}Teacher\\ Student\end{tabular} & \begin{tabular}[c]{@{}c@{}}ResNet32$\times$4\\ ResNet8$\times$4\end{tabular} & \begin{tabular}[c]{@{}c@{}}ResNet32$\times$4\\ ShuffleNetV2\end{tabular}  & \begin{tabular}[c]{@{}c@{}}VGG13\\ MobileNetV2\end{tabular} & \begin{tabular}[c]{@{}c@{}}ResNet50\\ VGG8\end{tabular}& \begin{tabular}[c]{@{}c@{}}VGG13\\ VGG8\end{tabular}  & ARI (\%) \\ \midrule
\begin{tabular}[c]{@{}c@{}}Teacher\\ Student\end{tabular} & \begin{tabular}[c]{@{}c@{}}57.92\\ 49.91 $\pm$ 0.16 \end{tabular}& \begin{tabular}[c]{@{}c@{}}57.92\\ 50.60 $\pm$ 0.23\end{tabular} & \begin{tabular}[c]{@{}c@{}}52.02\\ 44.20 $\pm$ 0.22\end{tabular}& \begin{tabular}[c]{@{}c@{}}55.44\\ 47.00 $\pm$ 0.17\end{tabular} & \begin{tabular}[c]{@{}c@{}}52.02\\ 47.00 $\pm$ 0.17\end{tabular}  & / \\ \midrule
KD      & 52.28 $\pm$ 0.07 & 57.27 $\pm$ 0.03        & 45.39  $\pm$ 0.59     & 51.50 $\pm$ 0.36 & 51.34 $\pm$ 0.08 & 123.18 \%\\
AT+KD   & 54.79 $\pm$ 0.23 & 57.56 $\pm$ 0.38      & 45.13  $\pm$ 0.60     & 51.42 $\pm$ 0.42  & 51.03 $\pm$ 0.28  &
122.61 \% \\
PKT+KD  & 54.11  $\pm$ 0.18& 58.33 $\pm$ 0.36        & 47.73  $\pm$ 0.31     & 51.45 $\pm$ 0.28  & 51.61 $\pm$ 0.28& 
35.51 \% \\
SP+KD   & 54.22  $\pm$ 0.41& 58.66 $\pm$ 0.25       & 48.10 $\pm$ 0.59      & 51.70 $\pm$ 0.12 & 51.51 $\pm$ 0.32  & 
29.98 \%\\
CC+KD   & 54.08  $\pm$ 0.32& 58.20 $\pm$ 0.06        & 47.67  $\pm$ 1.14     & 50.87 $\pm$ 0.20  & 51.07 $\pm$ 0.33& 
44.12 \%\\
RKD+KD  & 53.78  $\pm$ 0.15& 57.85 $\pm$ 0.24       & 48.10 $\pm$ 0.26      & 51.01 $\pm$ 0.23 & 50.59 $\pm$ 0.32  & 
46.70 \% \\
CRD+KD  & 55.53  $\pm$ 0.41& 58.95 $\pm$ 0.05        & 49.12 $\pm$ 0.04      & 52.87 $\pm$ 0.30  & 52.25 $\pm$ 0.26& 
7.88 \% \\ 
SSKD+KD & 55.10  $\pm$ 2.05& 57.48 $\pm$ 0.04        & 47.02 $\pm$ 0.90      & 52.36 $\pm$ 0.36 & 51.60 $\pm$ 0.16 &
35.51 \%\\ \midrule
HKD    & 55.53  $\pm$ 0.07 &58.83 $\pm$ 0.09          &49.53 $\pm$ 0.32       &52.20 $\pm$ 0.20   &51.97 $\pm$ 0.33& 
10.48 \%\\ 
HKD+KD    & \textbf{56.18 $\pm$ 0.12 } & \textbf{59.31 $\pm$ 0.01  }  & \textbf{49.57  $\pm$ 0.54}     & \textbf{53.30 $\pm$ 0.33  } & \textbf{52.62 $\pm$ 0.03  }& / \\ \bottomrule
    \end{tabular}
        \label{table:class2}
        \end{table*}

\begin{table*}[htpb]
\centering
\caption{Test accuracy (\%) of the student networks on the ImageNet dataset. The results of competing methods are obtained from \cite{chen2020cross}. }
\begin{tabular}{cccccccccc}
\hline
Method & Teacher & Student & KD    & FitNet & AT    & SP    & VID   & CRD   & HKD   \\ \hline
Top-1 Accuracy & 73.54   & 53.78   & 53.73 & 51.46  & 52.83 & 51.73 & 53.97 & 53.76 & \textbf{54.07} \\ \hline
\end{tabular}
\label{table:class3}
\end{table*}

\subsection{Baselines}
Several recently proposed knowledge distillation methods are compared, which can be categorized into two groups.
Their main difference is presented in Figure \ref{fig:intro}.

\textbf{(1) Individual Knowledge Distillation}: This group of methods capture knowledge contained in individual instances, including the logits in vanilla KD \cite{hinton2015distilling}, the attention map in AT \cite{zagoruyko2016paying}, and feature representation in CRD \cite{tian2019contrastive} and SSKD \cite{xu2020knowledge}.

\textbf{(2) Relational Knowledge Distillation}: This group of methods capture pairwise relational knowledge, including PKT \cite{passalis2018learning}, RKD\cite{park2019relational}, CCKD \cite{peng2019correlation}, SP\cite{tung2019similarity}.

We use the official implementation for these methods and follow the standard experimental settings.
For the SSKD method, we remove the data augmentation so that the training samples are consistent with other methods.

\subsection{Model Compression}
\textbf{Experimental Setup.} 
Model compression is one of the most fundamental applications of knowledge distillation.
The student network is learned by distilling knowledge from a fixed teacher network and ground truth labels.
We compare our method with several recent works with different teacher and student network architectures on CIFAR100, TinyImageNet and ImageNet datasets, as shown on Table \ref{table:class1}, Table \ref{table:class2} and Table \ref{table:class3} respectively.
All results are reported as the mean and variance of classification accuracy with five runs.
In order to obtain an intuitive sense about quantitative improvement, we adopt Average Relative Improvement (ARI) as the previous work \cite{tian2019contrastive}:
\begin{equation}
   \text{ARI}=\frac{1}{M} \sum_{i=1}^{M} \frac{\text{Acc}_{\text{HKD}}^{i}-\text{Acc}_{\text{BKD}}^{i}}{\text{Acc}_{\text{BKD}}^{i}-\text{Acc}_{\text{STU}}^{i}} \times 100 \%
\end{equation}
where M is the number of different architecture combinations and $\text{Acc}_{\text{HKD}}^{i},\text{Acc}_{\text{BKD}}^{i},\text{Acc}_{\text{STU}}^{i}$ refer to the accuracy of HKD, baseline knowledge distillation methods and regular trained student network.

\textbf{Results and Analysis.}
The basic observation is that our method outperforms the conventional student network and baseline methods on most teacher and student pairs.
Even without KD loss, our proposed HKD method still achieves comparable performance.
This demonstrates the effectiveness of HKD method in distilling holistic knowledge from the teacher network to guide the student network learning.

We also find that the existing relational knowledge distillation methods can not always outperform the individual knowledge distillation methods.
This implies that the noisy signal due to aligning all pairs of relations among instances may hurt the student network learning, motivating our KNN based graph construction for noise filtering.
Another interesting observation is that the HKD method is not restricted to the same teacher and student network architecture.
More surprisingly, we find that the HKD method sometimes gains slightly more improvement over conventional student network when the teacher and the student networks have different architectures.
For example, on the TinyImageNet dataset, when the teacher network is fixed to ResNet32$\times$4 architecture, the student gains 12.56 $\%$ improvement with ResNet8$\times$4 architecture.
However, 17.21 $\%$ improvement over conventional student network is gained when the student uses ShuffleNetV2.
When the student network is fixed with VGG8 architecture, 11.95 $\%$ improvement is gained when the teacher network uses VGG13 architecture.
However, 13.4 $\%$ improvement is gained when the teacher network uses ResNet50 architecture.
This demonstrates the advantage of utilizing mutual information to measure the teacher and student networks' alignment since it is not restricted to the same network architectures.

\begin{table}[]
\centering
\caption{Representation transferability experiments of the student network. The student network is trained on the CIFAR100 dataset and transferred to the TinyImageNet and the STL10 dataset. A linear classifier is evaluated on the frozen representations of the student network.}
\begin{tabular}{c|c|c}
    \hline
    Dataset       & TinyImageNet   & STL-10         \\ \hline
    T:ResNet50    & 30.79 $\pm$ 0.01 & 70.16 $\pm$ 0.07 \\ 
    S:MobileNetV2 & 23.01 $\pm$ 0.05     & 61.42 $\pm$ 0.10     \\ \hline
    KD            & 22.92 $\pm$ 0.13     & 61.25 $\pm$ 0.09     \\ 
    AT+KD         & 25.02 $\pm$ 0.01     & 62.05 $\pm$ 0.06               \\ 
    PKT+KD              & 26.04 $\pm$ 0.11               & 63.71 $\pm$ 0.05               \\ 
    SP+KD               & 24.98 $\pm$ 0.08               & 62.25 $\pm$ 0.13               \\ 
    CC+KD               & 25.68 $\pm$ 0.03               & 62.52 $\pm$ 0.10               \\ 
    RKD + KD              & 26.10 $\pm$ 0.03& 63.26 $\pm$ 0.03\\ 
    CRD + KD              & 28.98 $\pm$ 0.05               & 65.87 $\pm$ 0.10               \\ 
    SSKD + KD               & 24.24 $\pm$ 0.02               & 61.78 $\pm$ 0.02               \\ \hline
    HKD + KD              & \textbf{30.55 $\pm$ 0.03}               &  \textbf{67.28 $\pm$ 0.08}              \\ \hline
\end{tabular}
\label{table:transfer}
\end{table}

\subsection{Representation Transferability}
\textbf{Experimental Setup.}
To evaluate the transferability of representations learned by the student network, we follow the experiment setting of existing works \cite{tian2019contrastive,passalis2018learning,zagoruyko2016paying} and compare HKD with multiple baseline methods.
We first train the student network on the CIFAR100 dataset, and employ it to get representations of each data instance on the TinyImageNet and STL-10 datasets. Then, we froze these representations and evaluate the performance with a randomly initialized linear classifier to measure the student network's transferability.

\textbf{Results and Analysis.}
Table \ref{table:transfer} shows the experimental results of representation transferability from CIFAR100 dataset to TinyImageNet and STL10 datasets.
Among them, HKD achieves better performance on all the transferred datasets, which proves the transferability of representations learned by the HKD method.
We also observe that the conventional KD method performs worse than the student network.
 This indicates that only transferring the logits to the student network will limit the transferability of representations, motivating the HKD to transfer the holistic knowledge in a unified framework.

\subsection{Ablation Study}
To further show the benefit of distilling holistic knowledge, we design ablation studies on the CIFAR100 dataset.
We test both similar and different architectures for the teacher and the student networks.

\textbf{Graph Construction and Graph Neural Networks.}
In the HKD method, graph construction and graph neural networks play a critical role in defining holistic knowledge.
To explore the impact of different graph construction strategies, we test two graph construction strategies: random graph construction (Rand) and fully connected graph construction (FC).
To demonstrate graph neural networks' superiority in combining the graph topology and the instance features, we compare two basic graph-based representation learning strategies: sum-pooling (Sum) and mean-pooling (Mean).

\begin{figure}
    \centering
    \includegraphics[width=0.42\textwidth]{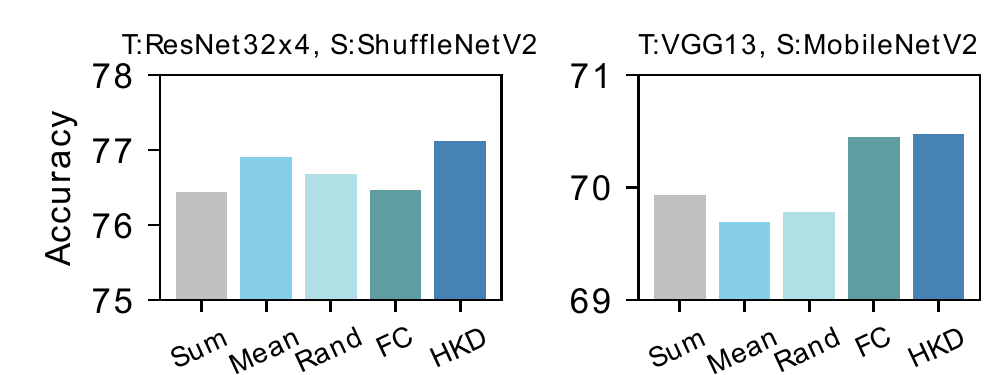}
    \caption{Ablation study on the definition of holistic knowledge for the HKD method on the CIFAR100 dataset.}
    \label{ablation:gnn}
\end{figure}

Figure \ref{ablation:gnn} illustrates the ablation study results.
We can observe that the HKD method that utilizes K-Nearest-Neighbors and graph neural networks achieves the best performance, demonstrating the effectiveness of the HKD method.

\begin{figure}
    \centering
    \includegraphics[width=0.42\textwidth]{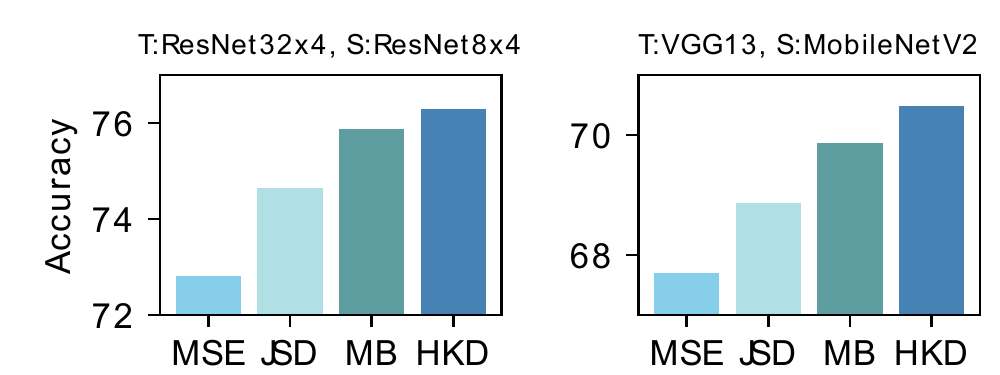}
    \caption{Ablation study on the training strategy for the HKD method on the CIFAR100 dataset.}
    \label{ablation:train}
\end{figure}

\textbf{Training Strategy.}
In the HKD method, we utilize mutual information with a graph-independent memory bank to guide holistic knowledge transfer.
To verify the advantage of such a training strategy, we compare with the following strategies:
the first one is the Mean Square Error (MSE) to measure the similarity between representations; the second one is the JS-divergence (JSD) with few negative samples in each mini-batch without memory bank; the third one is using memory bank (MB) to store the graph-based representations directly.

Figure \ref{ablation:train} illustrates the ablation study results.
We can observe the HKD method achieves better performance than the compared strategies, demonstrating the effectiveness of using the mutual information to measure the alignment, using the InfoNCE to estimate mutual information and memory bank for efficient training.

\begin{figure}
    \centering
    \includegraphics[width=0.42\textwidth]{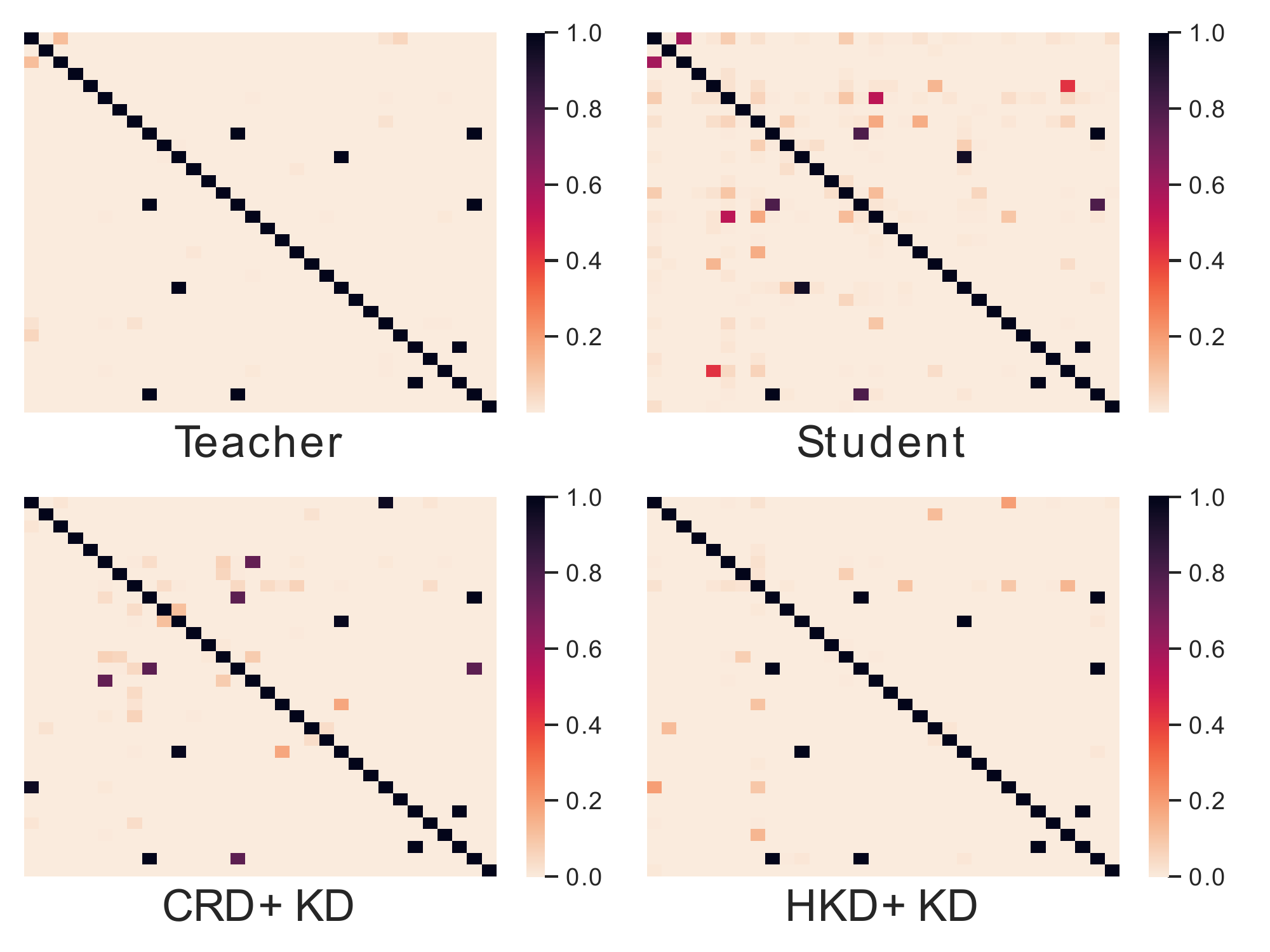}
    \caption{HeatMap visualization of four networks. The color denotes the strength of the similarity between pairs of instances.}
        \label{fig:vis}
\end{figure}

\subsection{Visualization and analysis}
\label{subsec:vis}
To delving into essence beyond results, we perform analysis based on visualization.
We first train a network and then randomly select a batch of data with 32 instances.
These instances are fed into four networks: the teacher network, the student network, CRD, and HKD.
We use cosine similarity to measure the pairwise similarity between the prediction and use different colors to represent the different strength of similarity.

Figure \ref{fig:vis} illustrates the experimental result.
Each block represents the pairwise cosine similarity between two instances. 
The darker color denotes higher cosine similarity while the lighter color denotes lower cosine similarity.
From this figure, we have the following observations:
First, most pairs have superficial similarities among the batch instances. This means most pairs of instances are not similar to each other, which motivates the HKD method of modeling the holistic knowledge instead of studying relationships between all pairs of instances.
Second, compare with the student network and the CRD network, our proposed HKD method have a more similar visualization result to the teacher network.
This demonstrates the effectiveness of the HKD method in distilling holistic knowledge from the teacher network.

\subsection{Hyperparameter Tuning}
In this subsection, we tune hyperparameters on the CIFAR100 dataset to test the sensitivity of the HKD method.
More specifically, we test the number of neighbors in K-Nearest-Neighbor and the $\beta$ in the loss function.

\begin{figure}
    \centering
    \includegraphics[width=0.42\textwidth]{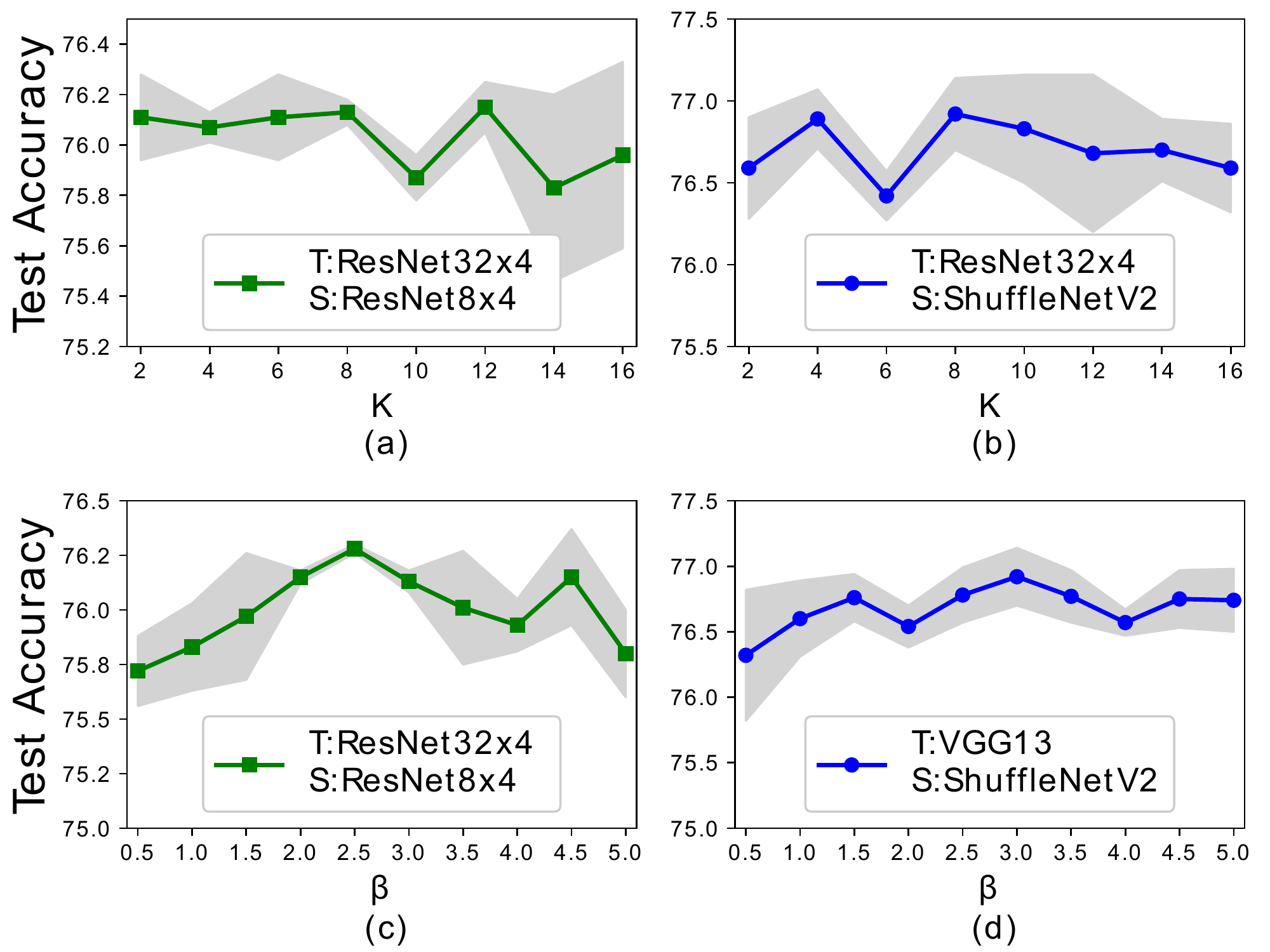}
    \caption{Hyper-parameter tuning of the HKD method. The first row of subfigures denotes the hyperparameter tuning results on the number of neighbors. The second row of subfigures denotes the hyperparameter tuning results on $\beta$.}
    \label{fig:hyper}
\end{figure}

Figure \ref{fig:hyper}-(a) and Figure \ref{fig:hyper}-(b) illustrate the impact of the number of nearest neighbors.
The basic observation is that HKD is not very sensitive to the number of neighbors in graph construction as the performance varies a little with different numbers of neighbors.
We get both the best performance in the two tested teacher network and student network architectures when we select $8$ neighborhood instances.
When the number of neighbors goes larger than $8$, we observe a decrease in performance, which is related to the over smoothing of graph neural networks.
Figure \ref{fig:hyper}-(c) and \ref{fig:hyper}-(d) illustrate the impact of $\beta$ on the HKD method.
We can observe that the HKD method slightly varies with different $\beta$.
This is reasonable as the holistic knowledge is of different importance with different $\beta$.

\section{Conclusion}
\label{sec:conclusion}
This paper proposes a holistic knowledge distillation method (HKD) with graph neural networks.
Compared with existing methods, the holistic knowledge integrates the individual and the relational knowledge while reserving their inherent correlations.
The graph neural networks (GNNs) are utilized to extract the holistic knowledge by aggregating feature representation from relational neighborhood samples.
The student network is trained under the supervision of holistic knowledge in a contrastive manner.
Extensive experiments are conducted on benchmark datasets to evaluate the performance and the motivation of HKD, the results demonstrate the effectiveness of the HKD
method.

\section{Acknowledgements}
This work is supported by the National Key Research and Development Program (Grant No. 2018YFB1403202), the National Natural Science Foundation of China (Grant No. 61972349,U1866602) Tsinghua GuoQiang Research Center (Grant 2020GQG1014) and Alibaba-Zhejiang University Joint Institute of Frontier Technologies.

{\small
\bibliographystyle{ieee_fullname}
\bibliography{main}

\begin{thebibliography}{10}\itemsep=-1pt

\bibitem{ba2014deep}
Jimmy Ba and Rich Caruana.
\newblock Do deep nets really need to be deep?
\newblock In {\em Advances in neural information processing systems}, pages
  2654--2662, 2014.

\bibitem{beck2018graph}
Daniel Beck, Gholamreza Haffari, and Trevor Cohn.
\newblock Graph-to-sequence learning using gated graph neural networks.
\newblock {\em arXiv preprint arXiv:1806.09835}, 2018.

\bibitem{chen2020online}
Defang Chen, Jian-Ping Mei, Can Wang, Yan Feng, and Chun Chen.
\newblock Online knowledge distillation with diverse peers.
\newblock In {\em AAAI}, pages 3430--3437, 2020.

\bibitem{chen2020cross}
Defang Chen, Jian-Ping Mei, Yuan Zhang, Can Wang, Zhe Wang, Yan Feng, and Chun
  Chen.
\newblock Cross-layer distillation with semantic calibration.
\newblock {\em arXiv preprint arXiv:2012.03236}, 2020.

\bibitem{chen2018modeling}
Jiawei Chen, Yan Feng, Martin Ester, Sheng Zhou, Chun Chen, and Can Wang.
\newblock Modeling users' exposure with social knowledge influence and
  consumption influence for recommendation.
\newblock In {\em Proceedings of the 27th ACM International Conference on
  Information and Knowledge Management}, pages 953--962, 2018.

\bibitem{chen2019samwalker}
Jiawei Chen, Can Wang, Sheng Zhou, Qihao Shi, Yan Feng, and Chun Chen.
\newblock Samwalker: Social recommendation with informative sampling strategy.
\newblock In {\em The World Wide Web Conference}, pages 228--239, 2019.

\bibitem{coates2011analysis}
Adam Coates, Andrew Ng, and Honglak Lee.
\newblock An analysis of single-layer networks in unsupervised feature
  learning.
\newblock In {\em Proceedings of the fourteenth international conference on
  artificial intelligence and statistics}, pages 215--223, 2011.

\bibitem{deng2009imagenet}
Jia Deng, Wei Dong, Richard Socher, Li-Jia Li, Kai Li, and Li Fei-Fei.
\newblock Imagenet: A large-scale hierarchical image database.
\newblock In {\em 2009 IEEE conference on computer vision and pattern
  recognition}, pages 248--255. Ieee, 2009.

\bibitem{devlin2019bert}
Jacob Devlin, Ming{-}Wei Chang, Kenton Lee, and Kristina Toutanova.
\newblock {BERT:} pre-training of deep bidirectional transformers for language
  understanding.
\newblock In {\em North American Chapter of the Association for Computational
  Linguistics: Human Language Technologies}, pages 4171--4186, 2019.

\bibitem{du2017topology}
Jian Du, Shanghang Zhang, Guanhang Wu, Jos{\'e}~MF Moura, and Soummya Kar.
\newblock Topology adaptive graph convolutional networks.
\newblock {\em arXiv preprint arXiv:1710.10370}, 2017.

\bibitem{garcia2017few}
Victor Garcia and Joan Bruna.
\newblock Few-shot learning with graph neural networks.
\newblock {\em arXiv preprint arXiv:1711.04043}, 2017.

\bibitem{hamilton2017inductive}
Will Hamilton, Zhitao Ying, and Jure Leskovec.
\newblock Inductive representation learning on large graphs.
\newblock In {\em Advances in neural information processing systems}, pages
  1024--1034, 2017.

\bibitem{he2016deep}
Kaiming He, Xiangyu Zhang, Shaoqing Ren, and Jian Sun.
\newblock Deep residual learning for image recognition.
\newblock In {\em Proceedings of the IEEE conference on computer vision and
  pattern recognition}, pages 770--778, 2016.

\bibitem{hinton2015distilling}
Geoffrey Hinton, Oriol Vinyals, and Jeff Dean.
\newblock Distilling the knowledge in a neural network.
\newblock {\em arXiv preprint arXiv:1503.02531}, 2015.

\bibitem{kipf2016semi}
Thomas~N Kipf and Max Welling.
\newblock Semi-supervised classification with graph convolutional networks.
\newblock {\em arXiv preprint arXiv:1609.02907}, 2016.

\bibitem{krizhevsky2009learning}
Alex Krizhevsky, Geoffrey Hinton, et~al.
\newblock Learning multiple layers of features from tiny images.
\newblock 2009.

\bibitem{lassance2020deep}
Carlos Lassance, Myriam Bontonou, Ghouthi~Boukli Hacene, Vincent Gripon, Jian
  Tang, and Antonio Ortega.
\newblock Deep geometric knowledge distillation with graphs.
\newblock In {\em ICASSP 2020-2020 IEEE International Conference on Acoustics,
  Speech and Signal Processing (ICASSP)}, pages 8484--8488. IEEE, 2020.

\bibitem{lee2018multi}
Chung-Wei Lee, Wei Fang, Chih-Kuan Yeh, and Yu-Chiang Frank~Wang.
\newblock Multi-label zero-shot learning with structured knowledge graphs.
\newblock In {\em Proceedings of the IEEE conference on computer vision and
  pattern recognition}, pages 1576--1585, 2018.

\bibitem{lee2019graph}
Seunghyun Lee and Byung~Cheol Song.
\newblock Graph-based knowledge distillation by multi-head attention network.
\newblock In {\em BMVC}, page 141, 2019.

\bibitem{li2020local}
Xiaojie Li, Jianlong Wu, Hongyu Fang, Yue Liao, Fei Wang, and Chen Qian.
\newblock Local correlation consistency for knowledge distillation.
\newblock In {\em European Conference on Computer Vision}, pages 18--33.
  Springer, 2020.

\bibitem{liu2019knowledge}
Yufan Liu, Jiajiong Cao, Bing Li, Chunfeng Yuan, Weiming Hu, Yangxi Li, and
  Yunqiang Duan.
\newblock Knowledge distillation via instance relationship graph.
\newblock In {\em Proceedings of the IEEE Conference on Computer Vision and
  Pattern Recognition}, pages 7096--7104, 2019.

\bibitem{oord2018representation}
Aaron van~den Oord, Yazhe Li, and Oriol Vinyals.
\newblock Representation learning with contrastive predictive coding.
\newblock {\em arXiv preprint arXiv:1807.03748}, 2018.

\bibitem{park2019relational}
Wonpyo Park, Dongju Kim, Yan Lu, and Minsu Cho.
\newblock Relational knowledge distillation.
\newblock In {\em Proceedings of the IEEE Conference on Computer Vision and
  Pattern Recognition}, pages 3967--3976, 2019.

\bibitem{passalis2018learning}
Nikolaos Passalis and Anastasios Tefas.
\newblock Learning deep representations with probabilistic knowledge transfer.
\newblock In {\em Proceedings of the European Conference on Computer Vision
  (ECCV)}, pages 268--284, 2018.

\bibitem{peng2019correlation}
Baoyun Peng, Xiao Jin, Jiaheng Liu, Dongsheng Li, Yichao Wu, Yu Liu, Shunfeng
  Zhou, and Zhaoning Zhang.
\newblock Correlation congruence for knowledge distillation.
\newblock In {\em Proceedings of the IEEE International Conference on Computer
  Vision}, pages 5007--5016, 2019.

\bibitem{rahimi2018semi}
Afshin Rahimi, Trevor Cohn, and Timothy Baldwin.
\newblock Semi-supervised user geolocation via graph convolutional networks.
\newblock {\em arXiv preprint arXiv:1804.08049}, 2018.

\bibitem{romero2014fitnets}
Adriana Romero, Nicolas Ballas, Samira~Ebrahimi Kahou, Antoine Chassang, Carlo
  Gatta, and Yoshua Bengio.
\newblock Fitnets: Hints for thin deep nets.
\newblock {\em arXiv preprint arXiv:1412.6550}, 2014.

\bibitem{sandler2018mobilenetv2}
Mark Sandler, Andrew Howard, Menglong Zhu, Andrey Zhmoginov, and Liang-Chieh
  Chen.
\newblock Mobilenetv2: Inverted residuals and linear bottlenecks.
\newblock In {\em Proceedings of the IEEE conference on computer vision and
  pattern recognition}, pages 4510--4520, 2018.

\bibitem{silver2017mastering}
David Silver, Julian Schrittwieser, Karen Simonyan, Ioannis Antonoglou, Aja
  Huang, Arthur Guez, Thomas Hubert, Lucas Baker, Matthew Lai, Adrian Bolton,
  et~al.
\newblock Mastering the game of go without human knowledge.
\newblock {\em Nature}, 550(7676):354--359, 2017.

\bibitem{simonyan2014very}
Karen Simonyan and Andrew Zisserman.
\newblock Very deep convolutional networks for large-scale image recognition.
\newblock {\em arXiv preprint arXiv:1409.1556}, 2014.

\bibitem{tian2019contrastive}
Yonglong Tian, Dilip Krishnan, and Phillip Isola.
\newblock Contrastive representation distillation.
\newblock {\em arXiv preprint arXiv:1910.10699}, 2019.

\bibitem{tschannen2019mutual}
Michael Tschannen, Josip Djolonga, Paul~K Rubenstein, Sylvain Gelly, and Mario
  Lucic.
\newblock On mutual information maximization for representation learning.
\newblock {\em arXiv preprint arXiv:1907.13625}, 2019.

\bibitem{tung2019similarity}
Frederick Tung and Greg Mori.
\newblock Similarity-preserving knowledge distillation.
\newblock In {\em Proceedings of the IEEE International Conference on Computer
  Vision}, pages 1365--1374, 2019.

\bibitem{wen2019preparing}
Tiancheng Wen, Shenqi Lai, and Xueming Qian.
\newblock Preparing lessons: Improve knowledge distillation with better
  supervision.
\newblock {\em arXiv preprint arXiv:1911.07471}, 2019.

\bibitem{wu2018unsupervised}
Zhirong Wu, Yuanjun Xiong, Stella~X Yu, and Dahua Lin.
\newblock Unsupervised feature learning via non-parametric instance
  discrimination.
\newblock In {\em Proceedings of the IEEE Conference on Computer Vision and
  Pattern Recognition}, pages 3733--3742, 2018.

\bibitem{xu2020knowledge}
Guodong Xu, Ziwei Liu, Xiaoxiao Li, and Chen~Change Loy.
\newblock Knowledge distillation meets self-supervision.
\newblock {\em arXiv preprint arXiv:2006.07114}, 2020.

\bibitem{yang2019training}
Chenglin Yang, Lingxi Xie, Siyuan Qiao, and Alan~L Yuille.
\newblock Training deep neural networks in generations: A more tolerant teacher
  educates better students.
\newblock In {\em Proceedings of the AAAI Conference on Artificial
  Intelligence}, volume~33, pages 5628--5635, 2019.

\bibitem{yang2019snapshot}
Chenglin Yang, Lingxi Xie, Chi Su, and Alan~L. Yuille.
\newblock Snapshot distillation: Teacher-student optimization in one
  generation.
\newblock In {\em Proceedings of the IEEE Conference on Computer Vision and
  Pattern Recognition}, pages 2859--2868, 2019.

\bibitem{zagoruyko2016paying}
Sergey Zagoruyko and Nikos Komodakis.
\newblock Paying more attention to attention: Improving the performance of
  convolutional neural networks via attention transfer.
\newblock {\em arXiv preprint arXiv:1612.03928}, 2016.

\bibitem{zhang2017active}
Mengni Zhang, Can Wang, Zhi Yu, Chao Shen, and Jiajun Bu.
\newblock Active learning for web accessibility evaluation.
\newblock In {\em Proceedings of the 14th International Web for All
  Conference}, pages 1--9, 2017.

\bibitem{zhang2018shufflenet}
Xiangyu Zhang, Xinyu Zhou, Mengxiao Lin, and Jian Sun.
\newblock Shufflenet: An extremely efficient convolutional neural network for
  mobile devices.
\newblock In {\em Proceedings of the IEEE conference on computer vision and
  pattern recognition}, pages 6848--6856, 2018.

\bibitem{zhou2018graph}
Jie Zhou, Ganqu Cui, Zhengyan Zhang, Cheng Yang, Zhiyuan Liu, Lifeng Wang,
  Changcheng Li, and Maosong Sun.
\newblock Graph neural networks: A review of methods and applications.
\newblock {\em arXiv preprint arXiv:1812.08434}, 2018.

\bibitem{zhou2020dge}
Sheng Zhou, Xin Wang, Jiajun Bu, Martin Ester, Pinggang Yu, Jiawei Chen, Qihao
  Shi, and Can Wang.
\newblock Dge: Deep generative network embedding based on commonality and
  individuality.
\newblock In {\em Proceedings of the AAAI Conference on Artificial
  Intelligence}, volume~34, pages 6949--6956, 2020.

\bibitem{zhou2018prre}
Sheng Zhou, Hongxia Yang, Xin Wang, Jiajun Bu, Martin Ester, Pinggang Yu,
  Jianwei Zhang, and Can Wang.
\newblock Prre: Personalized relation ranking embedding for attributed
  networks.
\newblock In {\em Proceedings of the 27th ACM International Conference on
  Information and Knowledge Management}, pages 823--832, 2018.

\end{thebibliography}
}

\section{Supplemental Material}

\subsection{Datasets and Architectures}
We conduct experiments on several benchmark datasets, including CIFAR100 \cite{krizhevsky2009learning},  STL-10 \cite{coates2011analysis}, TinyImageNet and ImageNet \cite{deng2009imagenet}.
Four architectures are used for the teacher and student networks, namely ResNet \cite{he2016deep},
VGG \cite{simonyan2014very}, ShuffleNet \cite{zhang2018shufflenet}, MobileNet \cite{sandler2018mobilenetv2}.

\begin{figure}
    \centering
    \includegraphics[width=0.48\textwidth]{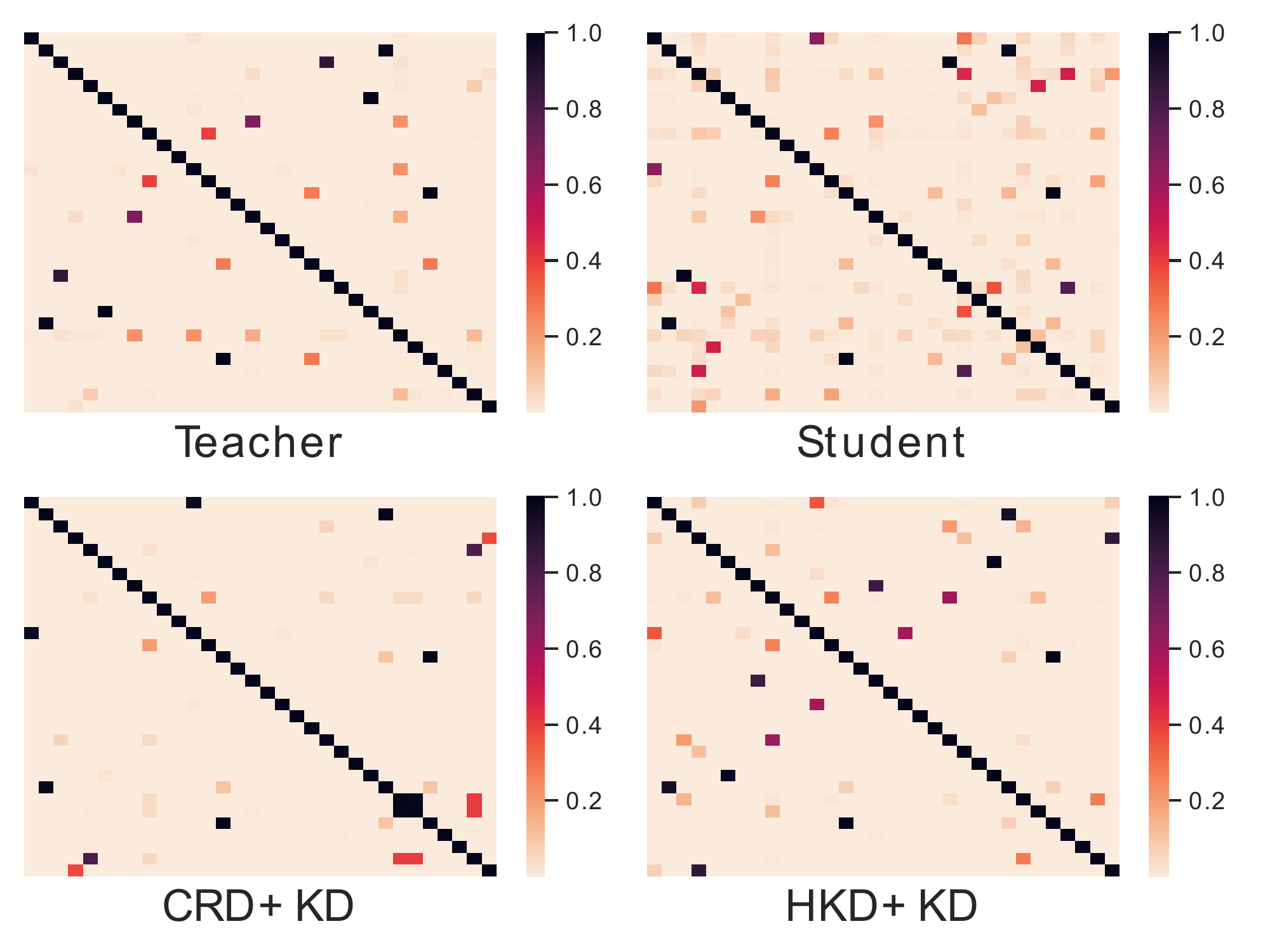}
    \caption{Visualization results on the CIFAR-100 datasets.}
    \label{vis1}
\end{figure}

\subsection{Implementation Details}
The CIFAR100 dataset consists of 50,000 images of size $32\times 32 $ with 500 images per class and 10,000 test images.
The TinyImageNet dataset is a subset of ImageNet, consisting of 100,000 images of size $64\times 64 $ from 200 classes.
STL-10 consists of 5000 labeled training images from 10 classes and 100,000 unlabeled images, and a test set of 8,000 images.
To keep our cross-modal transfer experiment's consistency, we down-sample each image to size $32\times 32 $.
We normalized all images by channel means and standard deviations.

Following the same experimental settings of existing works \cite{tian2019contrastive,park2019relational,peng2019correlation}, we use the SGD optimizer with momentum for all networks.
For MobileNetV2 and ShuffleNet, we use a learning rate of 0.01. For the rest of the networks, the learning rate is initialized with 0.05.
All the learning rates are decayed by 0.1 every 30 epochs after the first 150 epochs until the last 240 epoch.
We implement the networks and training procedures in Pytorch.

\subsection{Additional Visualization Results}
\begin{figure}
    \centering
    \includegraphics[width=0.48\textwidth]{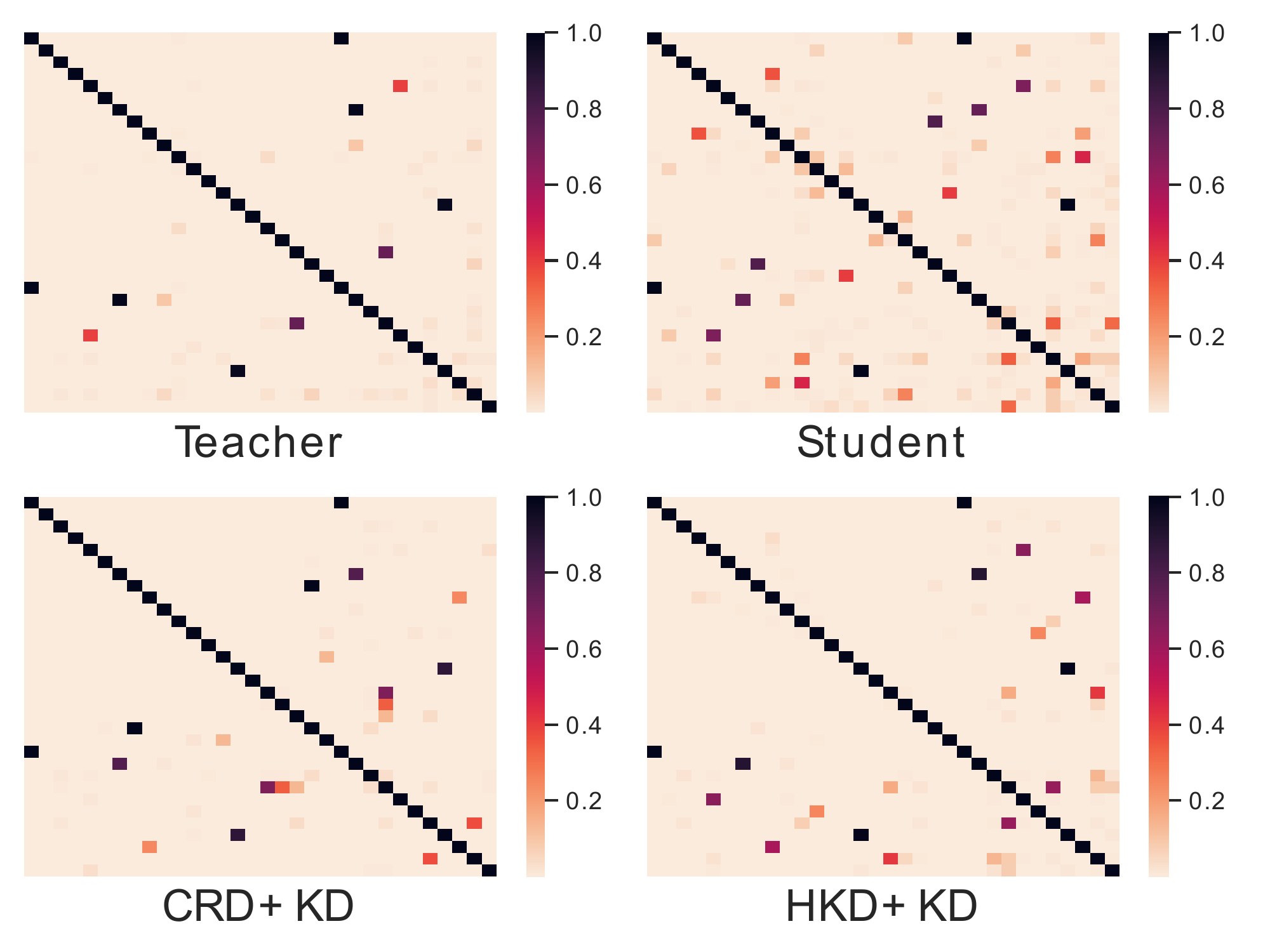}
    \caption{Visualization results on the CIFAR-100 datasets.}
    \label{vis2}
\end{figure}

We further provide more visualization results on the CIFAR-100 datasets, which is illustrated in Fig \ref{vis1} and Fig \ref{vis2}.
We observe the similar results that the proposed HKD method have similar topolopy structure with the teahcer network, which demonstrates the effectiveness of the proposed HKD method.

\subsection{Additional Parameter Analysis}
We further provide the parameter analysis on the number of layers in Graph Neural Networks.
More specifically, we test the one layer(L=1) and two layer(L=2) graph neural networks.
We do not set higher number of layers to avoid the over-smoothing problem.

Tabel \ref{tab:L} illustrates the experimental results on the CIFAR100 dataset, from which we can observe that the HKD method is not sensitive to the number of layers.

\begin{table}[htpb]
\caption{Parameter analysis on number of layers L}
\begin{tabular}{|c|c|c|c|}
\hline
Teacher & ResNet32x4 & VGG13       & ResNet50 \\ \hline
Student & ResNet8x4  & MobileNetV2 & VGG8     \\ \hline
L=1     & 76.13 $\pm$ 0.05 &70.48$\pm$0.25             &   74.85$\pm$0.26       \\ \hline
L=2     & 76.05$\pm$ 0.11  &70.28$\pm$0.07             &   74.82$\pm$0.24       \\ \hline
\end{tabular}
\label{tab:L}
\end{table}

\end{document}